\documentclass[twocolumn]{article}

\usepackage[english]{babel}

\usepackage[letterpaper,top=2cm,bottom=2cm,left=3cm,right=3cm,marginparwidth=1.75cm]{geometry}
\usepackage[english]{babel}
\usepackage{amsmath}
\usepackage{graphicx}
\usepackage{authblk}
\usepackage[colorlinks=true, allcolors=blue]{hyperref}
\usepackage[utf8]{inputenc}
\usepackage[english]{babel}
\usepackage{csquotes}
\usepackage{natbib}
\usepackage{balance}
\usepackage{array}
\usepackage{booktabs}
\usepackage{tabularx} 
\usepackage{multirow}
\usepackage{adjustbox}
\usepackage{multicol}

\title{Balancing the Scales: A Comprehensive Study on Tackling Class Imbalance in Binary Classification}

\author{
    Mohamed Abdelhamid and Abhyuday Desai \\
    \texttt{mohamed@readytensor.com, abhyuday.desai@readytensor.com}
}

\affil{Ready Tensor, Inc.}

\makeatletter
\def\@makefntext#1{%
  \parindent 1em%
  \noindent
  \hb@xt@1.8em{\hss\@makefnmark}#1}
\makeatother

\begin{document}

\twocolumn[
\date{}
  \maketitle
  \begin{abstract}
Class imbalance in binary classification tasks remains a significant challenge in machine learning, often resulting in poor performance on minority classes. This study comprehensively evaluates three widely-used strategies for handling class imbalance: Synthetic Minority Over-sampling Technique (SMOTE), Class Weights tuning, and Decision Threshold Calibration. We compare these methods against a baseline scenario of no-intervention across 15 diverse machine learning models and 30 datasets from various domains, conducting a total of 9,000 experiments. Performance was primarily assessed using the F1-score, although our study also tracked results on additional 9 metrics including F2-score, precision, recall, Brier-score, PR-AUC, and AUC. Our results indicate that all three strategies generally outperform the baseline, with Decision Threshold Calibration emerging as the most consistently effective technique. However, we observed substantial variability in the best-performing method across datasets, highlighting the importance of testing multiple approaches for specific problems. This study provides valuable insights for practitioners dealing with imbalanced datasets and emphasizes the need for dataset-specific analysis in evaluating class imbalance handling techniques.
\end{abstract}
  \vspace{1cm}
]

\section{Introduction}
Binary classification tasks frequently encounter imbalanced datasets, where one class significantly outnumbers the other. This imbalance can severely impact model performance, often resulting in classifiers that excel at identifying the majority class but perform poorly on the critical minority class. In fields such as fraud detection, disease diagnosis, and rare event prediction, this bias can have serious consequences.
To address this challenge, researchers have developed various techniques targeting different stages of the machine learning pipeline. These include Data preprocessing techniques, adjustments during model training, and Post-training calibration. In this study, we aim to provide a comprehensive comparison of three widely-used strategies for handling class imbalance:

\begin{itemize}
    \item Synthetic Minority Over-sampling Technique (SMOTE)

    \item Class Weights

    \item Decision Threshold Calibration

\end{itemize}

We compare these strategies with a Baseline approach (standard model training without addressing imbalance) to assess their effectiveness in improving model performance on imbalanced datasets. Our goal is to provide insights into which treatment methods offer the most significant improvements in various performance metrics such as F1-score, F2-score, accuracy, precision, recall, MCC, Brier score, Matthews Correlation Coefficient (MCC), PR-AUC, and AUC.
To ensure a comprehensive evaluation, this study encompasses:

\begin{itemize}
    \item 30 datasets from various domains, with sample sizes ranging from 500 to 20,000 and rare class percentages between 1\% and 15\%.

    \item 15 classifier models, including tree-based methods, boosting algorithms, neural networks, and traditional classifiers.

    \item Evaluation using 5-fold cross-validation.

\end{itemize}

In total, we conduct 9,000 experiments involving the 4 scenarios, 15 models, 30 datasets, and validation folds. This extensive approach allows us to compare these methods and their impact on model performance across a wide range of scenarios and algorithmic approaches, providing a robust foundation for understanding the effectiveness of different imbalance handling strategies in binary classification tasks.

\section{Related Works}

\subsection{Oversampling Techniques}
One of the most influential techniques developed to address class imbalance is the Synthetic Minority Over-sampling Technique (SMOTE), proposed by \cite{chawla2002smote}. SMOTE generates synthetic examples of the minority class by interpolating between existing samples. Since its introduction, the SMOTE paper has become one of the most cited in the field of imbalanced learning, with over 30,000 citations.
SMOTE's popularity has spurred the creation of many other oversampling techniques and numerous SMOTE variants. For example, \cite{kovacs2019empirical} documented 85 SMOTE-variants implemented in Python, including:
\begin{itemize}
    \item Borderline-SMOTE by \cite{han2005borderline}

    \item Safe-Level-SMOTE by \cite{bunkhumpornpat2009safe}

    \item SMOTE + Tomek and SMOTE + ENN by \cite{batista2004study}

\end{itemize}

While SMOTE has seen widespread use, its effectiveness has been called into question in high-dimensional settings. \cite{blagus2013smote} demonstrated that SMOTE's performance tends to degrade when applied to high-dimensional datasets. In these cases, synthetic examples generated by SMOTE may not be as informative, potentially introducing noise and reducing classifier performance.
\cite{elor2022smote} and \cite{van2007experimental} suggest the presence of better alternatives for handling class imbalance. This highlights the need for adaptations of SMOTE or alternative methods for handling imbalanced data in complex scenarios.

Furthermore, oversampling methods like SMOTE have been critiqued for potentially creating synthetic samples that do not reflect real-world distributions. \cite{hassanat2022stop} argue that these artificially generated instances can lead to overfitting, particularly when models are tested in real-world applications. This can result in models that perform well in controlled experimental conditions but fail to generalize effectively outside of those environments. Such concerns have prompted the development of more sophisticated techniques to address these challenges

 \subsection{Decision Threshold}
A critical factor in improving performance in imbalanced classification is the choice of evaluation metrics and thresholds. \cite{he2009learning} reviewed several methods for handling imbalanced datasets and stressed the importance of tailored metrics such as precision-recall curves and area under the ROC curve (AUC) for proper evaluation. Standard metrics like accuracy can be misleading when applied to imbalanced data, often inflating the model’s performance on the majority class while ignoring minority class errors.

In addition to using appropriate metrics, optimizing the classification threshold is crucial for imbalanced datasets. Many models use a default threshold of 0.5, which can be unsuitable for imbalanced data. \cite{zou2016finding} proposed a framework for finding the optimal classification threshold for imbalanced datasets, showing significant improvements in F-score by fine-tuning this threshold. Their method demonstrated that adjusting the threshold based on the dataset's characteristics helps to balance precision and recall, leading to better overall model performance.

\cite{leevy2023threshold} extended this work by exploring various threshold optimization techniques, concluding that optimal threshold selection without the use of Random Undersampling (RUS) often yields the best results. Their research highlighted the importance of balancing True Positive Rate (TPR) and True Negative Rate (TNR) in threshold-based optimization and showed that default thresholds often perform poorly on imbalanced datasets.

In a comparative study, \cite{hancock2022comparative} investigated several threshold optimization techniques for imbalanced datasets, including the use of metrics like precision, f-measure, Matthews Correlation Coefficient (MCC), and geometric mean of true positive rate (TPR) and true negative rate (TNR). Their results highlighted that optimized thresholds consistently outperformed the default threshold of 0.5, particularly when optimized with respect to the class distribution and specific performance metrics. This comparative approach underscores the importance of selecting appropriate thresholds based on the specific classification objectives and dataset characteristics.

\cite{hernandez2012unified} offered a comprehensive framework for understanding how performance metrics like AUC, precision, and accuracy relate to threshold choice. They demonstrated that the choice of threshold has a direct impact on expected classification loss, particularly when operating conditions (such as class distributions or misclassification costs) are uncertain. Their unified view of performance metrics highlights how different thresholding methods, such as fixed or score-driven thresholds, influence the overall effectiveness of models.

\cite{maloof2003learning} further explored the relationship between learning from imbalanced datasets and learning with unequal and unknown error costs. He argued that both problems could be approached similarly using ROC analysis to adjust decision thresholds or manipulate cost matrices. Maloof's study demonstrated that techniques like oversampling, undersampling, and threshold adjustment could generate similar results by producing classifiers that fall along the same ROC curve. This unified approach underscores the versatility of ROC analysis in addressing both class imbalance and cost-sensitive learning.

\subsection{Class Weights}

In addition to these techniques, adaptive weight optimization (AWO) has emerged as an effective method for addressing imbalanced datasets. \cite{huang2013adaptive} proposed an adaptive weighting approach that dynamically adjusts class weights to account for variations between the training and test sets. This method uses an evolutionary algorithm to optimize weight configuration, ensuring better classifier performance across both training and testing sets, even in highly imbalanced conditions. Their results demonstrated that adaptive weighting can outperform fixed weighting methods, particularly when the class imbalance is extreme.

Cost-sensitive learning (CSL) presents an alternative to oversampling by adjusting the cost associated with misclassifying the minority class. Instead of balancing the dataset through sample generation, CSL increases the weight of minority class errors during training. \cite{thai2010cost} combined CSL with resampling techniques, showing that optimizing the cost ratio as a hyperparameter can significantly improve performance on imbalanced datasets. By treating the cost ratio as a tunable parameter, this method provides a flexible solution that adapts to different imbalance levels.

Recent work has introduced advanced re-weighting strategies to improve imbalanced classification. \cite{guo2022learning} proposed a novel re-weighting method based on optimal transport (OT), which views the imbalanced training dataset as a distribution that needs to be aligned with a balanced meta-set. Their method optimizes the weights of training examples by minimizing the OT distance between the two distributions, achieving state-of-the-art performance across various tasks, including text, image, and point cloud classification(Learning to Re-weight). This approach represents a significant advancement in re-weighting techniques by decoupling the weight learning process from the classifier, providing a more robust solution for handling imbalanced datasets.

\subsection{Gaps in the Literature and Motivation for This Study}

While existing literature offers valuable insights into various techniques for handling class imbalance, there remains a notable gap in comprehensive, large-scale comparative studies that evaluate multiple methods across a wide range of datasets and models. Most prior works focus on specific techniques or limited sets of datasets, making it challenging to draw generalizable conclusions about the relative effectiveness of different approaches. Furthermore, there is a lack of studies that systematically examine how the performance of these techniques varies across different types of models and dataset characteristics. This gap in the literature motivated our study to provide a more holistic view of class imbalance handling strategies. By comparing SMOTE, Class Weights, and Decision Threshold Calibration across 30 diverse datasets and 15 different models, our work aims to offer practitioners and researchers a more comprehensive understanding of when and how to apply these techniques effectively. Additionally, our study's emphasis on dataset-level analysis addresses the need for more nuanced guidance in selecting appropriate methods for specific problem contexts.

\section{Methodology}

\subsection{Datasets}

We selected 30 datasets\footnote{\url{https://github.com/readytensor/rt-datasets-binary-class-imbalance}}
based on the following criteria:
\begin{itemize}
    \item Binary classification problems
    \item Imbalanced class distribution (minority class \(<\) 20\%)
    \item Sample size \(<=\) 20,000
    \item Feature count \(<=\) 100
    \item Real-world data from diverse domains
    \item Publicly available
\end{itemize}
The 20\% minority class threshold for class imbalance, while somewhat arbitrary, represents a reasonable cut-off point that is indicative of significant imbalance.
The limitations on sample size (\(<=\)) and feature count \(<=\) 100) were set to accommodate a wide range of real-world datasets while ensuring manageable computational resources for an experiment of our scale. This balance allows us to include diverse, practically relevant datasets without compromising the breadth of our study.
The focus on diverse domains ensures that our
models are tested across a wide range of industries and data characteristics, enhancing the generalizability of our findings.

\subsection{Models}

Our study employed a diverse set of 15 classifier models, encompassing a wide spectrum of algorithmic approaches and complexities. This selection ranges from simple baselines to advanced ensemble methods and neural networks, including tree-based models and various boosting algorithms. The diversity in our model selection allows us to assess how different imbalanced data handling techniques perform across various model types and complexities.

The following is a list of the models used in our experiments:

\begin{multicols}{2}
\begin{enumerate}
    \item AdaBoost
    \item Bagging
    \item CatBoost
    \item Decision Tree
    \item Explainable Boosting Machine
    \item Extra Trees
    \item FIGS (PiML)
    \item Gradient Boosting
    \item LightGBM
    \item Logistic Regression
    \item Relu DNN
    \item Random Forest
    \item Simple ANN
    \item SVM
    \item XGBoost
\end{enumerate}
\end{multicols}

A key consideration in our model selection process was ensuring that all four scenarios (Baseline, SMOTE, Class Weights, and Decision Threshold Calibration) could be applied consistently to each model. This criterion influenced our choices, leading to the exclusion of certain algorithms such as k-Nearest Neighbors (KNN) and Naive Bayes Classifiers, which do not inherently support the application of class weights. This careful selection process allowed us to maintain consistency across all scenarios while still representing a broad spectrum of machine learning approaches.

To ensure a fair comparison, we used the same preprocessing pipeline for all 15 models and scenarios. This pipeline includes steps such as one-hot encoding, standard scaling, and missing data imputation. The only difference in preprocessing occurs in the SMOTE scenario, where synthetic minority class examples are generated. Otherwise, the preprocessing steps are identical across all models and scenarios, ensuring that the only difference is the algorithm and the specific imbalance handling technique applied.

Additionally, each model's hyperparameters were kept constant across the Baseline, SMOTE, Class Weights, and Decision Threshold scenarios to ensure fair comparisons. The details of the hyperparameters used for each model can be found in the Appendix.

\subsection{Evaluation Metrics}
To comprehensively evaluate the performance of the models across different imbalanced data handling techniques, we tracked the following 10 metrics:
Accuracy, Precision, Recall, F1-Score, F2-Score, Matthews Correlation Coefficient (MCC), Log-Loss, AUC-Score, PR-AUC and Brier Score

Our primary focus is on the F1-score, a label metric that uses predicted classes rather than underlying probabilities. The F1-score provides a balanced measure of precision and recall, making it particularly useful for assessing performance on imbalanced datasets.

While real-world applications often employ domain-specific cost matrices to create custom metrics, our study spans 30 diverse datasets. The F1-score allows us to evaluate all four scenarios, including decision threshold tuning, consistently across this varied set of problems.

\subsection{Experimental Procedure}

Our experimental procedure was designed to ensure a robust and comprehensive evaluation of the four imbalance handling scenarios across diverse datasets and models. The process utilized a form of nested cross-validation for each dataset to ensure robust model evaluation and proper hyperparameter tuning. For the outer loop, we employed 5-fold cross-validation, where each dataset was split into five folds. Results were reported for all five test splits, providing mean and standard deviation values across the folds. Within this outer loop, for scenarios requiring hyperparameter tuning (SMOTE, Class Weights, and Decision Threshold Calibration), we implemented an inner validation process. This involved further dividing the training split from the outer loop into a 90\% train and 10\% validation split. The validation split was used exclusively for tuning hyperparameters. This nested approach allowed us to maintain the integrity of our test data while still allowing for proper hyperparameter optimization, ensuring a fair and thorough evaluation of each imbalance handling technique across a wide range of datasets and models.

\subsubsection{Scenario Descriptions}
We evaluated four distinct scenarios for handling class imbalance:

\begin{enumerate}
\item Baseline: This scenario involves standard model training without any specific treatment for class imbalance. It serves as a control for comparing the effectiveness of the other strategies.

\item SMOTE (Synthetic Minority Over-sampling Technique): In this scenario, we apply SMOTE to the training data to generate synthetic examples of the minority class.

\item Class Weights: This approach involves adjusting the importance of classes during model training, focusing on the minority class weight while keeping the majority class weight at 1.

\item Decision Threshold Calibration: In this scenario, we adjust the classification threshold post-training to optimize the model's performance on imbalanced data.
\end{enumerate}

Each scenario implements only one treatment method in isolation. We do not combine treatments across scenarios. Specifically:

\begin{itemize}
\item For scenarios 1, 2, and 3, we apply the default decision threshold of 0.5.
\item For scenarios 1, 2, and 4, the class weights are set to 1.0 for both positive and negative classes.
\item SMOTE is applied only in scenario 2, class weight adjustment only in scenario 3, and decision threshold calibration only in scenario 4.
\end{itemize}

\subsubsection{Hyperparameter Tuning}
For scenarios requiring hyperparameter tuning (SMOTE, Class Weights, and Decision Threshold), we employed a simple grid search strategy to maximize the F1-score measured on the single validation split (10\% of the training data) for each fold.
The grid search details for the three treatment scenarios were as follows:

\begin{itemize}

    \item SMOTE \newline
    We tuned the number of neighbors hyperparameter, performing a simple grid search over `k` values of 1, 3, 5, 7, and 9.

    \item Class Weights \newline
    In this scenario, we adjusted the class weights to handle class imbalance during model training. The tuning process involved adjusting the weight for the minority class relative to the majority class. If both classes were given equal weights (e.g., 1 and 1), no class imbalance handling was applied—this corresponds to the baseline scenario. For the balanced scenario, we set the minority class weight proportional to the class imbalance (e.g., if the majority/minority class ratio was 5:1, the weight for the minority class would be 5). We conducted grid search on the following factors: 0 (baseline case), 0.25, 0.5, 0.75, 1.0 (balanced), and 1.25 (over-correction). The optimal weight was selected based on the F1-score on the validation split.

    \item Decision Threshold Calibration \newline
    We tuned the threshold parameter from 0.05 to 0.5 with a step size of 0.05, allowing for a wide range of potential decision boundaries.

\end{itemize}

\textbf{Overall Scope of Experiments} \newline
This study contains 9,000 experiments driven by the following factors: 30 datasets, 15 models, 4 scenarios and 5-fold cross-validation. 

For each experiment, we recorded the 10 performance metrics across the five test splits.

\section{Results}

This section presents a comprehensive analysis of our experiments comparing four strategies for handling class imbalance in binary classification tasks. We begin with an overall comparison of the four scenarios (Baseline, SMOTE, Class Weights, and Decision Threshold Calibration) across all ten evaluation metrics. Following this, we focus on the F1-score metric to examine performance across the 15 classifier models and 30 datasets used in our study. The results are available in our GitHub repository.\footnote{\url{https://github.com/readytensor/rt-binary-class-imbalance-results}}

Our analysis is structured as follows:

\begin{enumerate}
    \item Overall performance comparison by scenario and metric
    \item Model-specific performance on F1-score
    \item Dataset-specific performance on F1-score
    \item Statistical analysis, including repeated measures tests and post-hoc pairwise comparisons
\end{enumerate}

\subsection{Overall Comparison} 

Table 2 presents the mean performance and standard deviation for all 10 evaluation metrics across the four scenarios: Baseline, SMOTE, Class Weights, and Decision Threshold Calibration.

\begin{table*}[h!]
\centering
\begin{tabular}{|l|c|c|c|c|}
\hline
\textbf{Metric} & \textbf{Baseline} & \textbf{SMOTE} & \textbf{Class Weights} & \textbf{Decision Threshold} \\ \hline
F1-Score &  $0.556 \pm 0.006$ & $0.605 \pm 0.006$ & $0.594 \pm 0.006$ & $\mathbf{\textcolor{blue}{0.617 \pm 0.005}}$\\ \hline
F2-Score &  $0.53 \pm 0.008$ & $0.641 \pm 0.01$ & $0.597 \pm 0.008$ & $\mathbf{\textcolor{blue}{0.649 \pm 0.006}}$\\ \hline
MCC & $0.506 \pm 0.006$ & $\mathbf{\textcolor{blue}{0.532 \pm 0.005}}$ & $0.525 \pm 0.006$ & $\mathbf{\textcolor{blue}{0.532 \pm 0.009}}$\\ \hline
Recall & $0.517 \pm 0.008$ & $0.683 \pm 0.014$ & $0.609 \pm 0.01$ & $\mathbf{\textcolor{blue}{0.693 \pm 0.01}}$\\ \hline
Precision & $\mathbf{\textcolor{blue}{0.676 \pm 0.004}}$ & $0.575 \pm 0.005$ & $0.636 \pm 0.004$ & $0.597 \pm 0.005$\\ \hline
PR-AUC & $\mathbf{\textcolor{blue}{0.638 \pm 0.008}}$ & $0.629 \pm 0.01$ & $0.637 \pm 0.006$ & $0.635 \pm 0.007$\\ \hline
AUC & $0.865 \pm 0.005$ & $0.864 \pm 0.005$ & $\mathbf{\textcolor{blue}{0.866 \pm 0.005}}$ & $0.862 \pm 0.006$\\ \hline
Accuracy & $\mathbf{\textcolor{blue}{0.938 \pm 0.001}}$ & $0.904 \pm 0.003$ & $0.922 \pm 0.002$ & $0.898 \pm 0.005$\\ \hline
Log-Loss & $\mathbf{\textcolor{blue}{0.452 \pm 0.009}}$ & $0.518 \pm 0.008$ & $0.442 \pm 0.01$ & $0.432 \pm 0.012$\\ \hline
Brier Score & $\mathbf{\textcolor{blue}{0.058 \pm 0.001}}$ & $0.078 \pm 0.001$ & $0.066 \pm 0.001$ & $\mathbf{\textcolor{blue}{0.058 \pm 0.001}}$\\ \hline
\end{tabular}
\caption{Mean performance and standard deviation of evaluation metrics across all scenarios. Best values per metric are in $\mathbf{\textcolor{blue}{bold, blue}}$ font.
}
\label{tab:results}
\end{table*}

The results represent aggregated performance across all 15 models and 30 datasets, providing a comprehensive overview of the effectiveness of each scenario in handling class imbalance. \newline

The results show that all three class imbalance handling techniques outperform the Baseline scenario in terms of F1-score:
\begin{enumerate}
    \item Decision Threshold Calibration achieved the highest mean F1-score (0.617 ± 0.005)
    \item SMOTE followed closely (0.605 ± 0.006)
    \item Class Weights showed improvement over Baseline (0.594 ± 0.006)
    \item Baseline had the lowest F1-score (0.556 ± 0.006)

\end{enumerate}

While our analysis primarily focuses on the F1-score, it's worth noting observations from the other metrics:

\begin{itemize}
    \item F2-score and Recall: Decision Threshold Calibration and SMOTE showed the highest performance, indicating these methods are particularly effective at improving the model's ability to identify the minority class.
\item Precision: The Baseline scenario achieved the highest precision, suggesting a more conservative approach in predicting the minority class.
\item MCC (Matthews Correlation Coefficient): SMOTE and Decision Threshold Calibration tied for the best performance, indicating a good balance between true and false positives and negatives.
\item PR-AUC and AUC: These metrics showed relatively small differences across scenarios. Notably, SMOTE and Class Weights did not deteriorate performance on these metrics compared to the Baseline. As expected, Decision Threshold Calibration, being a post-model adjustment, does not materially impact these probability-based metrics (as well as Brier-Score).
\item Accuracy: The Baseline scenario achieved the highest accuracy, which is common in imbalanced datasets where high accuracy can be achieved despite poor minority class detection.
\item Log-Loss: The Baseline scenario performed best, suggesting it produces the most well-calibrated probabilities. SMOTE showed the highest log-loss, indicating potential issues with probability calibration.
\item Brier-Score: As expected, the Baseline and Decision Threshold scenarios show identical performance, as Decision Threshold Calibration is a post-prediction adjustment and doesn't affect the underlying probabilities used in the Brier Score calculation. Notably, SMOTE performed significantly worse on this metric, indicating it produces poorly calibrated probabilities compared to the other scenarios.
\end{itemize}

Based on these observations, Decision Threshold Calibration demonstrates strong performance across several key metrics, particularly those focused on minority class prediction (F1-score, F2-score, and Recall). It achieves this without compromising the calibration of probabilities of the baseline model, as evidenced by the identical Brier Score. In contrast, while SMOTE improves minority class detection, it leads to the least well-calibrated probabilities, as shown by its poor Brier Score. This suggests that Decision Threshold Calibration could be particularly effective in scenarios where accurate identification of the minority class is crucial, while still maintaining the probability calibration of the original model.

\subsection{Results by model}

\begin{table*}[h!]
\centering
\begin{tabular}{|l|c|c|c|c|}
\hline
\textbf{Model} & \textbf{Baseline} & \textbf{SMOTE} & \textbf{Class Weights} & \textbf{Decision Threshold} \\ \hline
Logistic Regression & $0.452 \pm 0.003$ & $0.513 \pm 0.002$ & $0.557 \pm 0.012$ & $\textcolor{blue}{\mathbf{0.563 \pm 0.007}}$ \\ \hline
Decision Tree & $\textcolor{blue}{\mathbf{0.586 \pm 0.008}}$ & $0.584 \pm 0.008$ & $0.582 \pm 0.007$ & $0.582 \pm 0.008$ \\ \hline
FIGS & $0.525 \pm 0.008$ & $0.565 \pm 0.006$ & $\textcolor{blue}{\mathbf{0.590 \pm 0.007}}$ & $0.562 \pm 0.013$ \\ \hline
SVM                 & $0.529 \pm 0.013$ & $\textcolor{blue}{\mathbf{0.604 \pm 0.007}}$ & $0.537 \pm 0.010$ & $0.591 \pm 0.007$ \\ \hline
ReLU-DNN            & $0.534 \pm 0.013$ & $0.611 \pm 0.012$ & $0.532 \pm 0.009$ & $\textcolor{blue}{\mathbf{0.618 \pm 0.005}}$ \\ \hline
Simple ANN          & $0.575 \pm 0.011$ & $0.617 \pm 0.01$ & $0.570 \pm 0.007$ & $\textcolor{blue}{\mathbf{0.622 \pm 0.011}}$ \\ \hline
AdaBoost            & $0.438 \pm 0.006$ & $0.544 \pm 0.005$ & $\textcolor{blue}{\mathbf{0.585 \pm 0.010}}$ & $0.577 \pm 0.008$ \\ \hline
Extra Trees         & $0.566 \pm 0.012$ & $0.626 \pm 0.010$ & $0.560 \pm 0.011$ & $\textcolor{blue}{\mathbf{0.628 \pm 0.013}}$ \\ \hline
Bagging Classifier  & $0.568 \pm 0.008$ & $0.617 \pm 0.009$ & $0.618 \pm 0.005$ & $\textcolor{blue}{\mathbf{0.635 \pm 0.009}}$ \\ \hline
Random Forest       & $0.576 \pm 0.008$ & $0.632 \pm 0.007$ & $0.573 \pm 0.010$ & $\textcolor{blue}{\mathbf{0.643 \pm 0.013}}$ \\ \hline
Gradient Boosting   & $0.613 \pm 0.011$ & $0.628 \pm 0.006$ & $0.631 \pm 0.008$ & $\textcolor{blue}{\mathbf{0.639 \pm 0.008}}$ \\ \hline
EBM                 & $0.570 \pm 0.010$ & $0.626 \pm 0.008$ & $\textcolor{blue}{\mathbf{0.651 \pm 0.012}}$ & $0.650 \pm 0.009$ \\ \hline
XGBoost             & $0.619 \pm 0.012$ & $0.633 \pm 0.008$ & $0.627 \pm 0.012$ & $\textcolor{blue}{\mathbf{0.637 \pm 0.010}}$ \\ \hline
LightGBM            & $0.616 \pm 0.011$ & $0.640 \pm 0.013$ & $0.639 \pm 0.004$ & $\textcolor{blue}{\mathbf{0.650 \pm 0.006}}$ \\ \hline
CatBoost            & $0.568 \pm 0.008$ & $0.638 \pm 0.009$ & $0.651 \pm 0.006$ & $\textcolor{blue}{\mathbf{0.653 \pm 0.009}}$ \\ \hline
\end{tabular}
\caption{Mean F1-scores and standard deviations for each model across the four scenarios. Highest values per model are in bold, blue font.}
\label{tab:results}
\end{table*}
Key observations from these results include:

\begin{enumerate}
    \item Scenario Comparison: For each model, we compared the performance of the four scenarios (Baseline, SMOTE, Class Weights, and Decision Threshold Calibration). This within-model comparison is more relevant than comparing different models to each other, given the diverse nature of the classifier techniques.
    \item Decision Threshold Performance: The Decision Threshold Calibration scenario achieved the highest mean F1-score in 10 out of 15 models. Notably, even when it wasn't the top performer, it consistently remained very close to the best scenario for that model.
    \item Other Scenarios: Within individual models, Class Weights performed best in 3 cases, while SMOTE and Baseline each led in 1 case.
    \item Consistent Improvement: All three imbalance handling techniques generally showed improvement over the Baseline scenario across most models, with 1 exception.
\end{enumerate}

These results indicate Decision Threshold Calibration was most frequently the top performer across the 15 models. This suggests that post-model adjustments to the decision threshold is a robust strategy for improving model performance across different classifier techniques. However, the strong performance of other techniques in some cases underscores the importance of testing multiple approaches when dealing with imbalanced datasets in practice.

\subsection{Results by Dataset}
\begin{table*}[h!]
\centering
\begin{tabular}{|l|c|c|c|c|}
\hline
\textbf{Dataset} & \textbf{Baseline} & \textbf{SMOTE} & \textbf{Class Weights} & \textbf{Decision Threshold} \\ \hline
abalone binarized & $0.092 \pm 0.018$ & $0.358 \pm 0.016$ & $0.255 \pm 0.009$ & $\textcolor{blue}{\mathbf{0.377 \pm 0.017}}$ \\ \hline

auction & $0.74 \pm 0.019$ & $\textcolor{blue}{\mathbf{0.798 \pm 0.013}}$ & $0.786 \pm 0.02$ & $0.783 \pm 0.012$ \\ \hline

car eval binarized & $0.929 \pm 0.028$ & $0.932 \pm 0.024$ & $\textcolor{blue}{\mathbf{0.935 \pm 0.023}}$ & $0.916 \pm 0.027$ \\ \hline

chess & $0.834 \pm 0.036$ & $\textcolor{blue}{\mathbf{0.875 \pm 0.042}}$ & $0.843 \pm 0.04$ & $0.849 \pm 0.045$ \\ \hline

climate simulation crashes & $\textcolor{blue}{\mathbf{0.965 \pm 0.002}}$ & $0.96 \pm 0.006$ & $0.961 \pm 0.002$ & $0.96 \pm 0.002$ \\ \hline

club loan & $0.074 \pm 0.012$ & $0.245 \pm 0.004$ & $0.204 \pm 0.008$ & $\textcolor{blue}{\mathbf{0.313 \pm 0.008}}$ \\ \hline

coil 2000 & $0.07 \pm 0.009$ & $0.146 \pm 0.013$ & $0.157 \pm 0.014$ & $\textcolor{blue}{\mathbf{0.192 \pm 0.008}}$ \\ \hline

graduation & $0.053 \pm 0.012$ & $\textcolor{blue}{\mathbf{0.182 \pm 0.026}}$ & $0.112 \pm 0.023$ & $0.159 \pm 0.035$ \\ \hline

jm1 & $0.266 \pm 0.02$ & $0.405 \pm 0.019$ & $0.376 \pm 0.016$ & $\textcolor{blue}{\mathbf{0.422 \pm 0.013}}$ \\ \hline

kc1 & $0.361 \pm 0.035$ & $\textcolor{blue}{\mathbf{0.431 \pm 0.027}}$ & $0.401 \pm 0.022$ & $0.427 \pm 0.028$ \\ \hline

letter img & $0.905 \pm 0.013$ & $0.887 \pm 0.007$ & $\textcolor{blue}{\mathbf{0.915 \pm 0.01}}$ & $0.907 \pm 0.012$ \\ \hline

mammography & $0.637 \pm 0.074$ & $0.559 \pm 0.03$ & $0.631 \pm 0.052$ & $\textcolor{blue}{\mathbf{0.643 \pm 0.055}}$ \\ \hline

optical digits & $0.898 \pm 0.017$ & $0.907 \pm 0.015$ & $0.907 \pm 0.017$ & $\textcolor{blue}{\mathbf{0.909 \pm 0.013}}$ \\ \hline

ozone level & $0.331 \pm 0.058$ & $0.44 \pm 0.034$ & $0.37 \pm 0.055$ & $\textcolor{blue}{\mathbf{0.445 \pm 0.046}}$ \\ \hline

page blocks & $0.844 \pm 0.013$ & $0.824 \pm 0.023$ & $\textcolor{blue}{\mathbf{0.848 \pm 0.015}}$ & $0.844 \pm 0.016$ \\ \hline

pc1 & $0.321 \pm 0.12$ & $\textcolor{blue}{\mathbf{0.405 \pm 0.084}}$ & $0.355 \pm 0.077$ & $0.372 \pm 0.071$ \\ \hline

pen digits & $0.938 \pm 0.005$ & $0.941 \pm 0.003$ & $\textcolor{blue}{\mathbf{0.953 \pm 0.003}}$ & $0.948 \pm 0.004$ \\ \hline

pie chart & $0.199 \pm 0.044$ & $\textcolor{blue}{\mathbf{0.377 \pm 0.057}}$ & $0.276 \pm 0.065$ & $0.367 \pm 0.032$ \\ \hline

satellite & $\textcolor{blue}{\mathbf{0.729 \pm 0.076}}$ & $0.683 \pm 0.054$ & $0.705 \pm 0.069$ & $0.697 \pm 0.062$ \\ \hline

satimage & $0.593 \pm 0.019$ & $\textcolor{blue}{\mathbf{0.624 \pm 0.016}}$ & $0.62 \pm 0.018$ & $0.621 \pm 0.017$ \\ \hline

seismic bumps & $0.077 \pm 0.015$ & $0.204 \pm 0.021$ & $0.156 \pm 0.021$ & $\textcolor{blue}{\mathbf{0.234 \pm 0.02}}$ \\ \hline

sick euthyroid & $0.808 \pm 0.027$ & $0.804 \pm 0.019$ & $\textcolor{blue}{\mathbf{0.827 \pm 0.026}}$ & $0.824 \pm 0.029$ \\ \hline

solar flare & $0.08 \pm 0.062$ & $0.172 \pm 0.049$ & $0.158 \pm 0.031$ & $\textcolor{blue}{\mathbf{0.183 \pm 0.054}}$ \\ \hline

thoracic surgery & $0.906 \pm 0.006$ & $0.845 \pm 0.023$ & $0.904 \pm 0.006$ & $\textcolor{blue}{\mathbf{0.913 \pm 0.003}}$ \\ \hline

thyroid disease & $\textcolor{blue}{\mathbf{0.847 \pm 0.034}}$ & $0.835 \pm 0.017$ & $0.844 \pm 0.032$ & $0.842 \pm 0.034$ \\ \hline

us crime & $0.463 \pm 0.101$ & $\textcolor{blue}{\mathbf{0.487 \pm 0.038}}$ & $0.465 \pm 0.059$ & $0.485 \pm 0.051$ \\ \hline

vowel & $0.956 \pm 0.029$ & $0.946 \pm 0.023$ & $\textcolor{blue}{\mathbf{0.959 \pm 0.024}}$ & $0.948 \pm 0.028$ \\ \hline

wilt & $0.788 \pm 0.036$ & $0.82 \pm 0.027$ & $0.834 \pm 0.02$ & $\textcolor{blue}{\mathbf{0.838 \pm 0.017}}$ \\ \hline

wine quality & $0.23 \pm 0.034$ & $0.32 \pm 0.049$ & $0.303 \pm 0.033$ & $\textcolor{blue}{\mathbf{0.348 \pm 0.03}}$ \\ \hline

yeast & $0.738 \pm 0.055$ & $\textcolor{blue}{\mathbf{0.748 \pm 0.04}}$ & $0.748 \pm 0.046$ & $0.738 \pm 0.018$ \\ \hline
\end{tabular}
\caption{Mean F1-scores and standard deviations for each dataset across the four scenarios. Highest values per dataset are in bold, blue font.}
\label{tab:results}
\end{table*}

Key observations from these results include:
\begin{enumerate}
    \item Variability: There is substantial variability in which scenario performs best across different datasets, highlighting that there is no one-size-fits-all solution for handling class imbalance.
\item Scenario Performance:
\begin{itemize}
    \item Decision Threshold Calibration was best for 12 out of 30 datasets (40\%)
    \item SMOTE was best for 9 datasets (30\%)
    \item Class Weights was best for 7 datasets (23.3\%)
    \item Baseline was best for 3 datasets (10\%)
    \item There was one tie between SMOTE and Class Weights
\end{itemize}

\item Improvement Magnitude: The degree of improvement over the Baseline varies greatly across datasets, from no improvement to substantial gains (e.g., satellite vs abalone binarized).
\item Benefit of Imbalance Handling: While no single technique consistently outperformed others across all datasets, the three imbalance handling strategies generally showed improvement over the Baseline for most datasets.
\end{enumerate}

These results underscore the importance of testing multiple imbalance handling techniques for each specific dataset and task, rather than relying on a single approach. The variability observed suggests that the effectiveness of each method may depend on the unique characteristics of each dataset.

\subsection{Statistical Analysis}
To rigorously compare the performance of the four scenarios, we conducted statistical tests on the F1-scores aggregated by dataset (averaging across the 15 models for each dataset). \newline

\textbf{Repeated Measures ANOVA} \newline
We performed a repeated measures ANOVA to test for significant differences among the four scenarios. For this test, we have 30 datasets, each with four scenario F1-scores, resulting in 120 data points. The null hypothesis is that there are no significant differences among the mean F1-scores of the four scenarios. We use Repeated Measures ANOVA to account because we have multiple measurements (scenarios) for each dataset.

The test yielded a p-value of 2.01e-07, which is well below our alpha level of 0.05. This result indicates statistically significant differences among the mean F1-scores of the four scenarios. \newline

\textbf{Post-hoc Pairwise Comparisons} \newline
Following the significant ANOVA result, we conducted post-hoc pairwise comparisons using a Bonferroni correction to adjust for multiple comparisons. With 6 comparisons, our adjusted alpha level is 0.05/6 = 0.0083. \newline

\begin{table*}[h!]
\centering
\begin{tabular}{|l|c|c|c|}
\hline
\textbf{Scenario} & \textbf{Class Weights} & \textbf{Decision Threshold} & \textbf{SMOTE} \\ \hline
\textbf{Baseline} & \textcolor{blue}{$7.77 \times 10^{-5}$} & \textcolor{blue}{$2.26 \times 10^{-4}$} & \textcolor{blue}{$1.70 \times 10^{-3}$} \\ \hline
\textbf{Class Weights} & - & \textcolor{blue}{$2.06 \times 10^{-3}$} & $1.29 \times 10^{-1}$ \\ \hline
\textbf{Decision Threshold} & - & - & $2.83 \times 10^{-2}$ \\ \hline
\end{tabular}
\caption{P-values for pairwise comparisons. P-values below alpha level (Bonferroni-corrected) are in blue font.}
\label{tab:scenario_comparison}
\end{table*}

Key findings from the pairwise comparisons:

\begin{enumerate}
    \item The Baseline scenario is significantly different from all other scenarios (p $<$ 0.0083 for all comparisons).
    \item Class Weights is significantly different from Baseline and Decision Threshold, but not from SMOTE.
    \item There is no significant difference between SMOTE and Decision Threshold, or between SMOTE and Class Weights at the adjusted alpha level.

\end{enumerate}
These results suggest that while all three imbalance handling techniques (SMOTE, Class Weights, and Decision Threshold) significantly improve upon the Baseline, the differences among these techniques are less pronounced. The Decision Threshold approach shows a significant improvement over Baseline and Class Weights, but not over SMOTE, indicating that both Decision Threshold and SMOTE may be equally effective strategies for handling class imbalance in many cases.

\subsection{Discussion}
Our comprehensive study on handling class imbalance in binary classification tasks yielded several important insights:

\begin{enumerate}
    \item \textbf{Addressing Class Imbalance:} Our results strongly suggest that handling class imbalance is crucial for improving model performance. Across most datasets and models, at least one of the imbalance handling techniques outperformed the baseline scenario, often by a significant margin.

    \item \textbf{Effectiveness of SMOTE:} SMOTE demonstrated considerable effectiveness in minority class detection, showing significant improvements over the baseline in many cases. It was the best-performing method for 30\% of the datasets, indicating its value as a class imbalance handling technique. However, it's important to note that while SMOTE improved minority class detection, it also showed the worst performance in terms of probability calibration, as evidenced by its high Log-Loss and Brier Score. This suggests that while SMOTE can be effective for improving classification performance, it may lead to less reliable probability estimates. Therefore, its use should be carefully considered in applications where well-calibrated probabilities are crucial.

    \item \textbf{Optimal Method:} Decision Threshold Calibration emerged as the most consistently effective technique, performing best for 40\% of datasets and showing robust performance across different model types. It's also worth noting that among the three methods studied, Decision Threshold Calibration is the least computationally expensive. Given its robust performance and efficiency, it could be considered a strong default choice for practitioners dealing with imbalanced datasets.

    \item \textbf{Variability Across Datasets:} Despite the overall strong performance of Decision Threshold Calibration, we observed substantial variability in the best-performing method across datasets. This underscores the importance of testing multiple approaches for each specific problem.

    \item \textbf{Importance of Dataset-Level Analysis:} Unlike many comparative studies on class imbalance that report results at the model level aggregated across datasets, our study emphasizes the importance of dataset-level analysis. We found that the best method can vary significantly depending on the dataset characteristics. This observation highlights the necessity of analyzing and reporting findings at the dataset level to provide a more nuanced and practical understanding of imbalance handling techniques.

\end{enumerate}

\section{Study Limitations and Future Work}
While our study provides valuable insights, it's important to acknowledge its limitations:

\begin{enumerate}
    \item \textbf{Fixed Hyperparameters:} We used previously determined model hyperparameters. Future work could explore the impact of optimizing these hyperparameters specifically for imbalanced datasets. For instance, adjusting the maximum depth in tree models might allow for better modeling of rare classes.

    \item \textbf{Statistical Analysis:} Our analysis relied on repeated measures ANOVA and post-hoc tests. A more sophisticated approach, such as a mixed-effects model accounting for both dataset and model variability simultaneously, could provide additional insights and is an area for future research.

    \item \textbf{Dataset Characteristics:} While we observed variability in performance across datasets, we didn't deeply analyze how specific dataset characteristics (e.g., sample size, number of features, degree of imbalance) might influence the effectiveness of different methods. Future work could focus on identifying patterns in dataset characteristics that predict which imbalance handling technique is likely to perform best.

    \item \textbf{Limited Scope of Techniques:} Our study focused on three common techniques for handling imbalance. Future research could expand this to include other methods or combinations of methods.

    \item \textbf{Performance Metric Focus:} While we reported multiple metrics, our analysis primarily focused on F1-score. Different applications might prioritize other metrics, and the relative performance of these techniques could vary depending on the chosen metric.
\end{enumerate}
These limitations provide opportunities for future research to further refine our understanding of handling class imbalance in binary classification tasks. Despite these limitations, our study offers valuable guidance for practitioners and researchers dealing with imbalanced datasets, emphasizing the importance of addressing class imbalance and providing insights into the relative strengths of different approaches.

\section{Conclusion}
Our study provides a comprehensive evaluation of three widely used strategies—SMOTE, Class Weights, and Decision Threshold Calibration—for handling imbalanced datasets in binary classification tasks. Compared to a baseline scenario where no intervention was applied, all three methods demonstrated substantial improvements in key metrics related to minority class detection, particularly the F1-score, across a wide range of datasets and machine learning models.

The results show that addressing class imbalance is crucial for improving model performance. Decision Threshold Calibration emerged as the most consistent and effective technique, offering significant performance gains across various datasets and models. SMOTE also performed well, and Class Weights tuning proved to be a reasonable method for handling class imbalance, showing moderate improvements over the baseline.

However, the variability in performance across datasets highlights that no single method is universally superior. Therefore, practitioners should consider testing multiple approaches and tuning them based on their specific dataset characteristics.

While our study offers valuable insights, certain areas could be explored in future research. We fixed the hyperparameters across scenarios to ensure fair comparisons, holding all factors constant except for the treatment. Future research could investigate optimizing hyperparameters specifically for imbalanced datasets. Additionally, further work could explore how specific dataset characteristics influence the effectiveness of different techniques. Expanding the scope to include other imbalance handling methods or combinations of methods would also provide deeper insights. While our primary analysis focused on the F1-score, results for other metrics are available, allowing for further exploration and custom analyses based on different performance criteria.

In conclusion, our findings emphasize the importance of addressing class imbalance and offer guidance on choosing appropriate techniques based on dataset and model characteristics. Decision Threshold Calibration, with its strong and consistent performance, can serve as a valuable starting point for practitioners dealing with imbalanced datasets, but flexibility and experimentation remain key to achieving the best results.

\balance
\bibliographystyle{chicago}
\bibliography{references}

\newpage
\onecolumn
\appendix
\section*{Appendix: Model Hyperparameters}

The following models are implemented using the respective packages:
\begin{itemize}
    \item AdaBoostClassifier, BaggingClassifier, DecisionTreeClassifier, ExtraTreesClassifier, GradientBoostingClassifier, LogisticRegression, RandomForestClassifier, and SVM are implemented using the \texttt{scikit-learn} package.
    \item CatBoostClassifier is implemented using the \texttt{catboost} package.
    \item LightGBM is implemented using the \texttt{lightgbm} package.
    \item Explainable Boosting Machine is implemented using the \texttt{InterpretML} package.
    \item FIGS and ReluDNN are implemented using the \texttt{piml} package.
    \item Simple ANN is implemented using \texttt{PyTorch}.
    \item XGBoost is implemented using the \texttt{xgboost} package.
\end{itemize}
Hyperparameters that are not included are set to the default values provided by the respective packages. We are listing the most important hyperparameters that were reviewed and set specifically in our experimentation.

\begin{multicols}{2}
\textbf{AdaBoostClassifier}
\begin{itemize}
    \item estimator: Decision Tree
    \item n\_estimators: 100
    \item learning\_rate: 0.1
    \item algorithm: SAMME.R
\end{itemize}

\textbf{BaggingClassifier}
\begin{itemize}
    \item estimator: Decision Tree
    \item n\_estimators: 50
    \item max\_samples: 1
    \item max\_features: 1
    \item bootstrap: True
\end{itemize}

\textbf{CatboostClassifier}
\begin{itemize}
    \item iterations: 100
    \item learning\_rate: 0.1
    \item depth: 6
    \item l2\_leaf\_reg: 3
\end{itemize}

\textbf{DecisionTreeClassifier}
\begin{itemize}
    \item criterion: gini
    \item max\_depth: None
    \item min\_samples\_split: 2
    \item min\_samples\_leaf: 1
\end{itemize}

\subsection*{Explainable Boosting Machine}
\begin{itemize}
    \item min\_samples\_leaf: 2
    \item learning\_rate: 0.01
    \item max\_leaves: 3
\end{itemize}

\textbf{ExtraTreesClassifier}
\begin{itemize}
    \item n\_estimators: 100
    \item criterion: gini
    \item max\_depth: None
    \item min\_samples\_split: 2
    \item min\_samples\_leaf: 1
\end{itemize}

\textbf{FIGS}
\begin{itemize}
    \item max\_iter: 30
    \item max\_depth: 5
    \item min\_samples\_leaf: 2
    \item learning\_rate: 1
\end{itemize}

\textbf{Gradient Boosting}
\begin{itemize}
    \item max\_depth: 100
    \item learning\_rate: 0.2
    \item n\_estimators: 300
    \item min\_samples\_split: 8
    \item min\_samples\_leaf: 4
\end{itemize}

\textbf{LightGBM}
\begin{itemize}
    \item boosting\_type: gbdt
    \item n\_estimators: 150
    \item num\_leaves: 31
    \item learning\_rate: 0.1
    \item max\_depth: -1
\end{itemize}

\textbf{Logistic Regression}
\begin{itemize}
    \item penalty: elasticnet
    \item C: 0.5
    \item l1\_ratio: 0.5
    \item max\_iter: 100
    \item solver: lbfgs
\end{itemize}

\textbf{Random Forest}
\begin{itemize}
    \item n\_estimators: 100
    \item criterion: gini
    \item max\_depth: None
    \item min\_samples\_split: 2
    \item min\_samples\_leaf: 1
\end{itemize}

\textbf{ReluDNN}
\begin{itemize}
    \item hidden\_layer\_sizes: (40, 40)
    \item max\_epochs: 500
    \item learning\_rate: 0.001
\end{itemize}

\textbf{SVM}
\begin{itemize}
    \item C: 1
    \item kernel: rbf
    \item gamma: scale
\end{itemize}

\textbf{Simple ANN (PyTorch)}
\begin{itemize}
    \item lr: 0.001
    \item activation: tanh
    \item hidden\_layers: 2
\end{itemize}

\textbf{XGBoost}
\begin{itemize}
    \item booster: gbtree
    \item n\_estimators: 200
    \item eta: 0.5
    \item gamma: 0.05
    \item max\_depth: 10
\end{itemize}
\end{multicols}
\end{document}